\documentclass[runningheads]{llncs}
\usepackage{amsmath,amssymb} 
\usepackage[T1]{fontenc}
\usepackage{graphicx}
\usepackage{color}
\def\etal{{\em et al.~}}
\def \ie{{\em i.e.}}
\def \eg{{\em e.g.}}
\def \etc{{\em etc.~}}

\usepackage[linesnumbered,ruled]{algorithm2e}
\usepackage{amsfonts}
\usepackage{bm}
\usepackage{epsfig,latexsym}
\usepackage{float}
\usepackage{indentfirst}
\usepackage{times}
\usepackage{psfrag}
\usepackage{cite}
\usepackage{subfigure}
\usepackage{textcomp}
\usepackage{multirow}
\usepackage{multicol}
\usepackage{stfloats}
\usepackage{url}
\usepackage{booktabs}
\usepackage{threeparttable}
\usepackage{arydshln}
\usepackage{pifont}   
\usepackage{bbding}
\usepackage{rotating}
\usepackage{bbm}
\usepackage{colortbl}
\usepackage{array}
\definecolor{mygray}{gray}{.9}
\usepackage{hyperref}

\usepackage[normalem]{ulem}
\newcommand{\tabincell}[2]{\begin{tabular}{@{}#1@{}}#2\end{tabular}}

\begin{document}
\title{Cross-Architecture Knowledge Distillation}
%
%
\author{Yufan Liu\inst{1,2}\orcidID{0000-0002-8426-9335} \and
Jiajiong Cao\inst{5} \and
Bing Li \inst{1,4}\thanks{Corresponding author.}\and
Weiming Hu\inst{1,2,3} \and
Jingting Ding\inst{5} \and
Liang Li\inst{5}}
\authorrunning{Y. Liu et al.}
%
\institute{National Laboratory of Pattern Recognition, Institute of Automation, Chinese Academy of Sciences, Beijing, China \and
School of Artificial Intelligence, University of Chinese Academy of Sciences, Beijing, China    \and
CAS Center for Excellence in Brain Science and Intelligence Technology, Beijing, China   \and
PeopleAI, Inc., Beijing, China  \and
Ant Financial Service Group, Beijing, China
\\
\email{bli@nlpr.ia.ac.cn}
}

\maketitle              
\begin{abstract}
Transformer attracts much attention because of its ability to learn global relations and superior performance. In order to achieve higher performance, it is natural to distill complementary knowledge from Transformer to convolutional neural network (CNN). However, most existing knowledge distillation methods only consider homologous-architecture distillation, such as distilling knowledge from CNN to CNN. They may not be suitable when applying to cross-architecture scenarios, such as from Transformer to CNN. To deal with this problem, a novel cross-architecture knowledge distillation method is proposed. Specifically, instead of directly mimicking output/intermediate features of the teacher, partially cross attention projector and group-wise linear projector are introduced to align the student features with the teacher’s in two projected feature spaces. And a multi-view robust training scheme is further presented to improve the robustness and stability of the framework. Extensive experiments show that the proposed method outperforms 14 state-of-the-arts on both small-scale and large-scale datasets.

\end{abstract}
\section{Introduction}

Knowledge distillation (KD) has become a fundamental topic for model performance promotion. It has been successfully applied to various applications including model compression~\cite{cheng2017survey} and knowledge transfer~\cite{tan2018survey}. KD usually adopts a teacher-student framework, where the student model is trained under the guidance of the teacher's knowledge. The knowledge is usually defined by soft outputs or intermediate features of the teacher model.

\begin{figure}[t]
	\begin{center}
		\vspace{-.2em}
		\includegraphics[width=.96\linewidth]{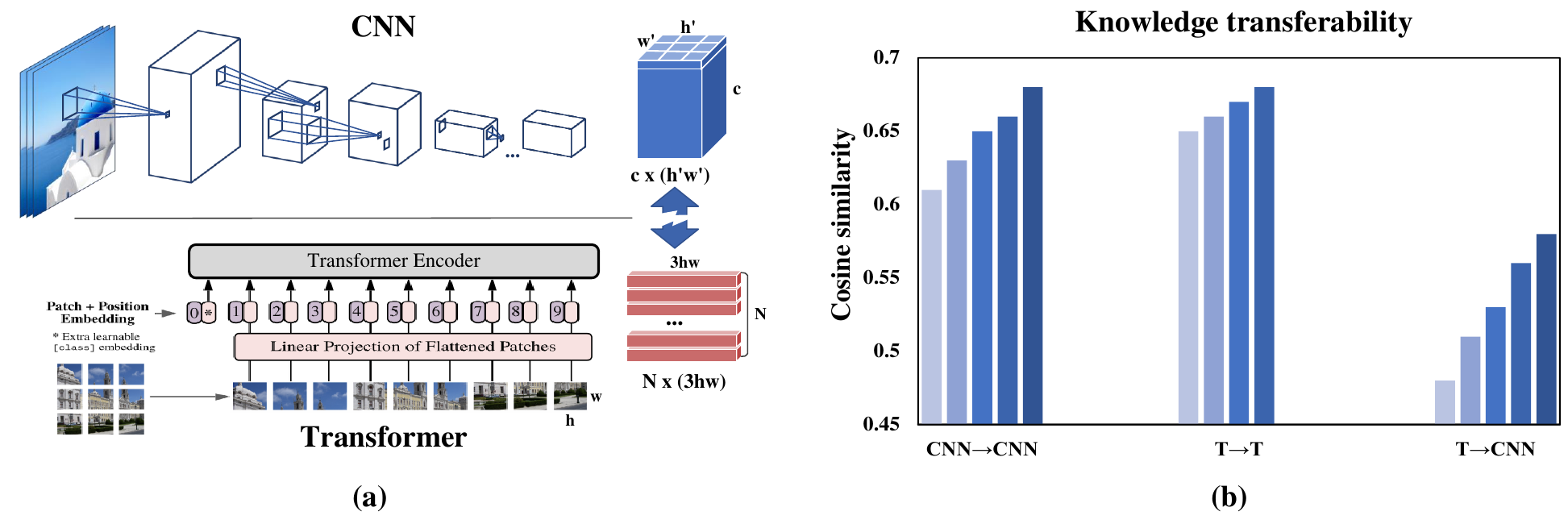}
	\end{center}
	\vspace{-2.2em}
	\caption{(a) The comparison of CNN and Transformer. The formation of the features are absolutely different. (b) The cosine similarity between features from different models on ImageNet. Note that the features are mapped into the same dimension by a linear projection. For ``CNN$\rightarrow$CNN'', the bars represent the similarities between CNN ResNet152 and CNNs \{ResNet18, ResNet32, ResNet50, ResNet101, ResNet152\}; For ``T$\rightarrow$T'', the bars represent the similarities between Transformer ViT-L/16 and Transformers \{ViT-B/32, ViT-B/16, ViT-L/32, ViT-L/16\}; For ``T$\rightarrow$CNN'', the bars represent the similarities between Transformer ViT-L/16 and CNNs \{ResNet18, ResNet32, ResNet50, ResNet101, ResNet152\}.}
	\vspace{-2.0em}
	\label{fig:fig1}
\end{figure}

Existing KD methods focus on convolutional neural network (CNN). However, there recently emerge many new networks such as Transformer. It shows superior on different computer vision tasks including image classification~\cite{dosovitskiy2020image} and detection~\cite{carion2020end}, while its huge computation and limited platform acceleration support limits the application of Transformer, especially for edge devices. On the other hand, with several years of development, there are sufficient acceleration libraries including CUDA~\cite{cuda_2007}, TensorRT~\cite{tensorrt_2022} and NCNN~\cite{ncnn_2017}, making CNN hardware friendly on both servers and edge devices. To this end, it is a natural idea to distill the knowledge from high-performance Transformer to compact CNN. However, there is a large gap between the two architectures. As shown in Figure \ref{fig:fig1}-(a), Transformer consists of self-attention-based transformer blocks while CNN contains a sequence of convolutional blocks. Further, the features are arranged in a totally different way. The intermediate outputs of CNNs are formed with $c$ channels of $h'\times w'$ feature maps. Different from CNN, the features of Transformer consist of $N$ feature vectors with $3hw$ elements, where $N$ refers to the patch number.

Unfortunately, existing methods focus on homologous-architecture KD such as CNN $\rightarrow$CNN and Transformer$\rightarrow$Transformer, which are not suitable for the cross-architecture scenarios. As shown Figure \ref{fig:fig1}-(b), the knowledge ``transferability'' is defined quantitatively. In particular, the output feature of the student is aligned to the feature space of the teacher, and then, the cosine similarity of the aligned student feature vector and the teacher feature vector is computed. For homologous-architecture cases, the transferability is between $0.6-0.7$, while it is much lower, typically lower than 0.55, on the cross-architecture condition. Consequently, it is more difficult to distill knowledge across different architectures and a new KD framework should be designed to deal with it.

In this work, a novel cross-architecture knowledge distillation method is proposed to bridge the large gap between Transformer and CNN.
With the help of the proposed framework, the knowledge from Transformer is efficiently transferred to the student CNN network and the knowledge transferability is significantly improved via this method. {It encourages the student to learn both local spatial features (with the original CNN model) and the complementary global features (from the transformer teacher	model).}
In particular, two projectors including a partially cross attention (PCA) projector and a group-wise linear (GL) projector, are designed. Instead of directly mimicking the output of the teacher, these two projectors align the intermediate student feature into two different feature spaces and knowledge distillation is further operated in the two feature spaces. The PCA projector maps the student feature into the Transformer attention space of the teacher. This projector encourages the student to learn the global relation from the Transformer teacher. The GL projector maps the student feature into the Transformer feature space in a pixel-by-pixel manner. This projector directly alleviates the feature formation differences between the teacher and the student. In addition, to alleviate the instability caused by the diversity in the cross-architecture framework, we propose a cross-view robust training scheme. Multi-view samples are generated to disturb the student network. And a multi-view adversarial discriminator is constructed to distinguish the teacher features and the disturbed student features, while the student is trained to confuse the discriminator. After convergence, the student can be more robust and stable.

Extensive experiments are conducted on both large-scale datasets and small-scale datasets, including ImageNet~\cite{ILSVRC15} and CIFAR~\cite{krizhevsky2009learning}. The experimental results of different teacher-student pairs demonstrate that the proposed method stably performs better than 14 state-of-the-arts. In summary, the main contributions of our work are three-fold: 
\begin{itemize}
	\item We propose a cross-architecture knowledge distillation framework to distill excellent Transformer knowledge to guide CNN. In this framework, partially cross attention (PCA) projector and group-wise linear (GL) projector are designed to align the student feature space and promote the transferability between teacher features and student features. 
	\item We propose a multi-view robust training scheme to improve the stability and robustness of the student network.
	\item Experimental results show that the proposed method is effective and outperforms 14 state-of-the-arts on both large-scale datasets and small-scale datasets.
\end{itemize}

\section{Related Work}
Hinton \etal \cite{hinton2015distilling} proposes the concept of knowledge distillation, using the soft output of teacher to guide the learning of student. Recently, it has been applied mainly to model compression \cite{cheng2017survey} and knowledge transfer \cite{tan2018survey}. Different formations of distilled knowledge are explored to better guide the student network, including final output~\cite{hinton2015distilling,ba2013deep} and hint layer knowledge~\cite{zagoruyko2016paying,romero2014fitnets,heo2019comprehensive,huang2017like,yim2017gift,liu2019knowledge,song2022spot,song2021tree}. For hint layer knowledge, many endeavors have been taken to match the student hint layers and the teacher-guided layers. For example, AT \cite{zagoruyko2016paying} defines single-channel attention maps as knowledge. However, the computation of the attention maps causes channel-dimension information loss. FitNet \cite{romero2014fitnets} directly distills the features from intermediate layers without information loss. However, this restriction is somewhat hard and not all the information is beneficial. 
Liu \etal \cite{liu2019knowledge} distill the knowledge called instance relationship graph (IRG), which contains instance
feature, instance feature relationship and feature space transformation. It is not limited by the dimension mismatch between the teacher and the student. 

The methods above all focus on convolutional neural network (CNN). Recently, Transformer becomes increasingly popular because of its impressive performance. However, due to the totally different architecture, many previous KD methods can not be directly applied to Transformers. There are some works \cite{touvron2021training,wang2020minilm,aguilar2020knowledge} studying knowledge distillation between Transformers. DeiT~\cite{touvron2021training} proposes a distillation token similar to the class token, to make the student Transformer learn the hard label from the teacher and ground truth (GT). MINILM \cite{wang2020minilm} focuses on the attention mechanisms in Transformer and distills the corresponding self-attention information. IR \cite{aguilar2020knowledge} distills the internal representations (\eg, self-attention map) from the teacher Transformer to the student Transformer.

In summary, existing methods usually present a transformation to match the teacher's features and the student's features. However, nearly all of them require similar or even the same architecture between teacher and student. 
To deal with the cross-architecture knowledge distillation problem, we carefully design projectors to match the teacher and the student in the same feature space. Consequently, a compact student CNN model can well learn the global feature from a teacher Transformer model despite the big gap in the architectures.

\section{The Proposed Method}
In this section, the framework of the proposed method is first introduced. Then, two key components of the framework including cross-architecture projectors and a cross-view robust training scheme are presented. The former is constructed to alleviate the feature mismatch for cross-architecture scenarios and help the student learn the global relation of the features, while the latter is adopted to improve the robustness and stability of the student. Finally, the loss function and training procedure are described. 

\subsection{Framework}
\begin{figure}[t]
	\begin{center}
		\vspace{-.6em}
		\includegraphics[width=1.0\linewidth]{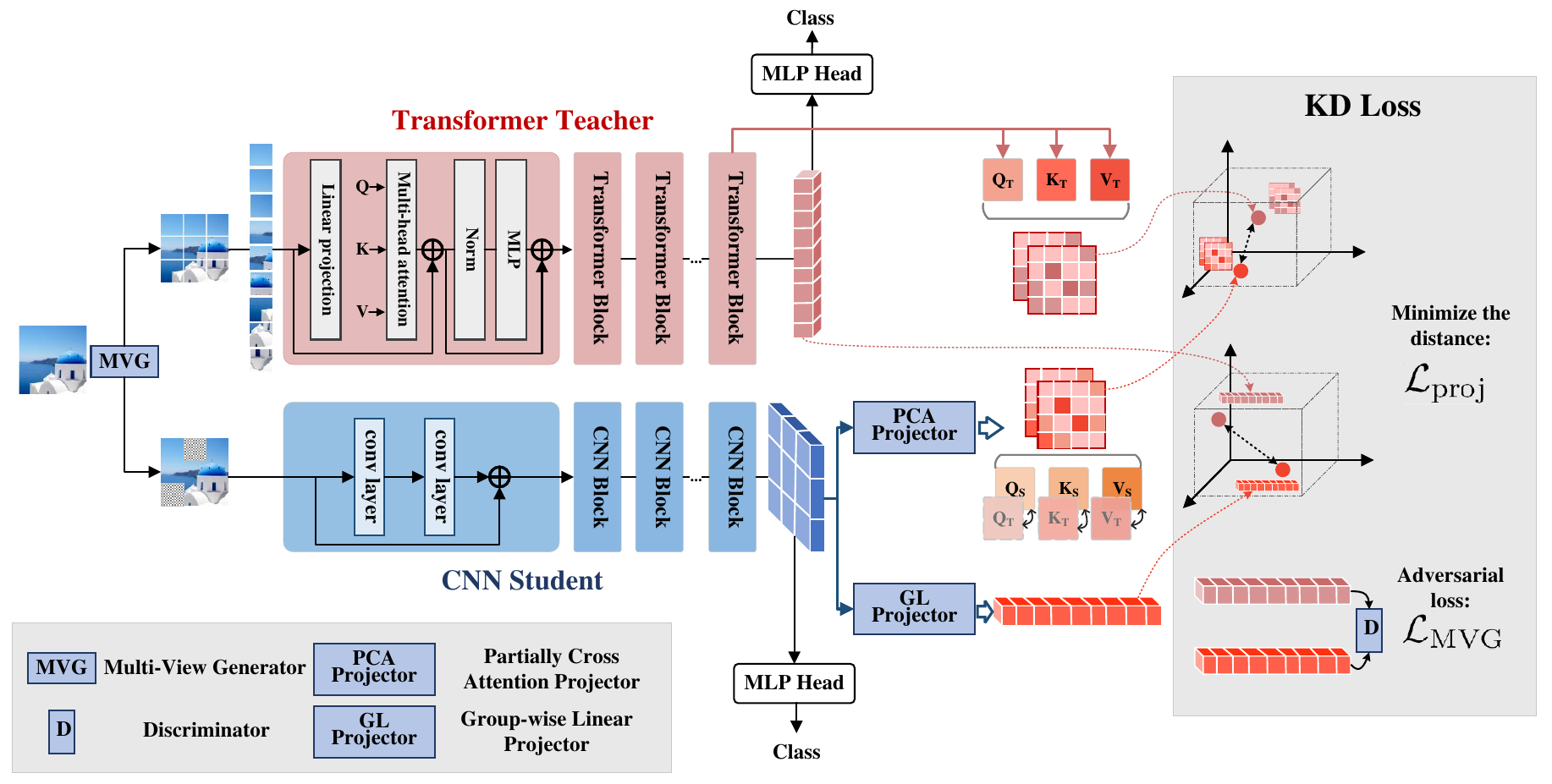}
	\end{center}
	\vspace{-2.0em}
	\caption{Overall framework of the proposed method.}
	\label{fig:framework}
	\vspace{-2.0em}
\end{figure}

The overall framework of the proposed method is depicted in Figure \ref{fig:framework}. In this figure, the upper pink network represents the teacher network, while the lower blue network is the student network. 
For the transformer teacher $\mathbf{\Theta}^{\mathrm{T}}$, the input sample $\mathbf{x}\in\mathbb{R}^{3\times H \times W}$ is divided into $(N=\frac{HW}{hw})$ patches $\{x_n\in\mathbb{R}^{3\times h \times w}\}_{n=1}^N$. After the inference of several transformer blocks, the feature $\mathbf{h}_{\mathrm{T}}\in\mathbb{R}^{N \times (3hw)}$ is generated. And the final predicted possibility is then computed via a multi layer perceptron (MLP) head as shown in Figure \ref{fig:framework}. 
For the CNN student $\mathbf{\Theta}^{\mathrm{S}}$, it receives the whole image without patch-wise partition as input. Similarly, after the inference of several CNN blocks, the final student feature $\mathbf{h}_{\mathrm{S}}\in\mathbb{R}^{c\times (h'w')}$ can be obtained. Note that $c$ is the channel number and $h'w'=\frac{HW}{2^{2s}}$. The $s$ denotes the number of CNN stages (usually equals 4). It is then used to predict the class. 

Due to the differences of the design principles and architectures between transformers and CNNs, it is hard to make the student features directly mimic the teacher features using the existing KD methods. To solve this problem, we propose a cross-architecture projector which consists of a partially cross attention (PCA) projector and a group-wise linear (GL) projector. The PCA projector maps the student features into the transformer attention space. By mapping the CNN feature space to this attention space, it is easier for the student to learn the global relationship among different regions by minimizing the distances between the student attention maps and the teacher attention maps. 
The GL projector maps the student features into the transformer feature space. In this transformer feature space, the student is guided to mimic the global transformer features in a pixel-by-pixel manner. 

To improve the robustness and stability of the student, a cross-view robust training scheme is proposed. Multi-view samples are generated by a multi-view generator which randomly conducts some transformations and generates mask and noise adding to the inputs. Fed with the multi-view inputs, the student generates different features. A multi-view adversarial discriminator is constructed to distinguish the teacher features and the student features in the transformer feature space. Then the goal is to puzzle the discriminator.

Eventually,  we integrate the proposed losses and give end-to-end training to obtain a strong student network.

\subsection{Cross-architecture projector}

\subsubsection{(1) Partially cross attention projector}
Partially cross attention (PCA) projector maps the student feature space into transformer attention space. It is designed to map the CNN features to Query, Key, Value matrices and then mimic the attention mechanism. It consists of three $3\times 3$ convolutional layers:
\begin{flalign}
	\begin{aligned}
		\label{eqn:PCA}
		\{Q_{\mathrm{S}}, K_{\mathrm{S}}, V_{\mathrm{S}}\} = 
		\mathrm{Proj_1}(\mathbf{h}_{\mathrm{S}}),
	\end{aligned}
\end{flalign}
where the matrixes $Q_{\mathrm{S}}, K_{\mathrm{S}}, V_{\mathrm{S}}$ are computed and aligned to mimic the query $Q_{\mathrm{T}}$, the key $K_{\mathrm{T}}$ and the value $V_{\mathrm{T}}$ of the Transformer teacher. In the transformer attention space, the self-attention of the student is calculated as:
\begin{flalign}
	\begin{aligned}
		\label{eqn:self_attn}
		\mathrm{Attn_S} = softmax(\dfrac{Q_{\mathrm{S}}(K_{\mathrm{S}})^{T}}{\sqrt{d}})V_{\mathrm{S}},
	\end{aligned}
\end{flalign}
in which $d$ is the query size. The calculation of $\mathrm{Attn_T}$ is similar. Hence, we can minimize the distance between the attention maps of the teacher and those of the student to guide the student network. To further improve the robustness of the student, we construct the partially cross attention of the student to replace the original $\mathrm{Attn_S}$:
\begin{equation}
	\begin{aligned}
		\label{eqn:PCattn}
		&\mathrm{PCAttn_S} = softmax(\dfrac{g(Q_{\mathrm{S}})(g(K_{\mathrm{S}}))^{T}}{\sqrt{d}})g(V_{\mathrm{S}}), \\
		&\mathrm{s.t.} \quad g(M_{\mathrm{S}}(i,j)) = 
		\left\{\begin{aligned}
			&M_{\mathrm{T}}(i,j), \quad p\ge 0.5 \\
			&M_{\mathrm{S}}(i,j), \quad p\ \textless\  0.5 \\
		\end{aligned}	\right.    
		, (M=Q, K, V).
	\end{aligned}
\end{equation}
Note that $(i,j)$ denotes the matrix element index of $M$. The function $g(\cdot)$ replaces the $Q_{\mathrm{S}}, K_{\mathrm{S}}, V_{\mathrm{S}}$ matrixes of the student by the corresponding matrixes of the teacher, with the probability $p$ subject to uniform distribution. 
In this manner, the loss is constructed:
\begin{flalign}
	\begin{aligned}
		\label{eqn:loss1}
		\mathcal{L}_{\mathrm {proj_1}} = ||\mathrm{Attn_T} - \mathrm{PCAttn_S}||_2^2 
		+ ||\frac{V_{\mathrm{T}} \cdot V_{\mathrm{T}}}{\sqrt{d}} - \frac{V_{\mathrm{S}} \cdot V_{\mathrm{S}}}{\sqrt{d}} ||_2^2,
	\end{aligned}
\end{flalign}
to make the student mimic the teacher in the attention space.

\subsubsection{(2) Group-wise linear projector}
Group-wise linear (GL) projector maps the student feature into transformer feature space. It consists of several shared-weight fully-connected (FC) layers:
\begin{flalign}
	\begin{aligned}
		\label{eqn:GL}
		\mathbf{h'}_{\mathrm{S}} = 
		\mathrm{Proj_2}(\mathbf{h}_{\mathrm{S}}),
	\end{aligned}
\end{flalign}
where $\mathbf{h'}_{\mathrm{S}}\in\mathbb{R}^{N \times (3hw)}$ is aligned to have the same dimension with teacher feature $\mathbf{h}_{\mathrm{T}}$. Specifically, for the regular image input with size of $224\times 224$, the dimensions are $\mathbf{h}_{\mathrm{S}}\in\mathbb{R}^{256 \times 196}$ and $\mathbf{h'}_{\mathrm{S}}\in\mathbb{R}^{196 \times 768}$. In order to realize a pixel-by-pixel mapping manner, the projector needs at least 196 FC layers with $256\times 768$ parameters. each of them maps the pixel from the original feature space to the corresponding ``pixel'' in the transformer space. A large number of FC layers may cause huge computation.  
In order to obtain a compact projector, we propose the \textbf{group-wise} linear projector where a $4\times 4$ neighborhood shares an FC layer. Hence, the GL projector only contains 16 FC layers. Furthermore, \textit{drop-out} is also adopted to reduce the computation and improve the robustness. 
Finally, after obtaining the new aligned student feature, the loss is computed as:
\begin{flalign}
	\begin{aligned}
		\label{eqn:loss2}
		\mathcal{L}_{\mathrm {proj_2}} = ||\mathbf{h}_{\mathrm{T}} - \mathbf{h'}_{\mathrm{S}}||_2^2,
	\end{aligned}
\end{flalign}
to minimize the distance between the teacher feature and the student feature in the transformer feature space. 

\subsection{Cross-view robust training}
Due to the big difference between the architectures of the teacher and the student, it is not that easy for the student to learn to be robust. To improve the robustness and the stability of the student network, we proposed a cross-view robust training scheme. The proposed training scheme contains two important components, \ie, a multi-view generator (MVG) and the corresponding multi-view adversarial discriminator. 
The MVG takes the original image as the input, and generates images with different transformations with some probability: 
\begin{equation}
	\begin{aligned}
		\label{eqn:MVG}
		\tilde{\mathbf{x}} = \mathrm{MVG}(\mathbf{x}) = 
		\left\{\begin{aligned}
			&\mathrm{Trans}(\mathbf{x}), \quad p\ge 0.5 \\
			&\mathbf{x}, \quad p\ \textless\  0.5 \\
		\end{aligned}	\right., 
	\end{aligned}
\end{equation}
in which $\mathrm{Trans}(\cdot)$ contains the common transformations, such as color jettering, random crop, rotation, patch-wise mask, \etc The probability $p$ is also subject to the uniform distribution. 
These transformed versions of the samples are then fed to the student network. 
Subsequently, the multi-view adversarial discriminator is constructed to distinguish the teacher feature $\mathbf{h}_{\mathrm{T}}$ and the transformed student feature $\mathbf{h'}_{\mathrm{S}}$, which is comprised of a three-FC-layer network. In this manner, the target of the cross-view robust training is to confuse the discriminator and obtain a robust student feature. The training loss of the discriminator is computed as: 
\begin{flalign}
	\begin{aligned}
		\label{eqn:D_loss}
		\mathcal{L}_{\mathrm{MAD}} = \frac{1}{m}\sum_{k=1}^{m}\left[-\log D(\mathbf{h}_{\mathrm{T}}^{(k)}) - \log(1-D(\mathbf{h'}_{\mathrm{S}}^{(k)}))\right].
	\end{aligned}
\end{flalign}
Note that $D(\cdot)$ denotes the multi-view adversarial discriminator. And $m$ is the total number of training samples. 
For the student network which can be seen as the generator in the adversarial training, the loss is written as: 
\begin{flalign}
	\begin{aligned}
		\label{eqn:G_loss}
		\mathcal{L}_{\mathrm{MVG}} = \frac{1}{m}\sum_{k=1}^{m}\left[ \log(1-D(\mathbf{h'}_{\mathrm{S}}^{(k)}))\right].
	\end{aligned}
\end{flalign}
Minimizing this loss can help to generate the student feature $\mathbf{h'}_{\mathrm{S}}$ which distributes similarly to that of the teacher feature $\mathbf{h}_{\mathrm{T}}$.

\subsection{Optimization}
In this subsection, we introduce the overall optimization and the training procedure of the proposed method. In order to train the student network, the loss function can be obtained by: 
\begin{flalign}
	\begin{aligned}
		\label{eqn:G_loss}
		\mathcal{L}_{\mathrm{total}} = (\mathcal{L}_{\mathrm{proj_1}} + \mathcal{L}_{\mathrm{proj_2}}) + \lambda\cdot\mathcal{L}_{\mathrm{MVG}},
	\end{aligned}
\end{flalign}
in which $\lambda$ is the penalty coefficient balancing the loss terms. 
For the multi-view adversarial discriminator, the loss function is $\mathcal{L}_{\mathrm{MAD}}$ in Equation (\ref{eqn:D_loss}). 

The overall training procedure of the proposed method is summarized in Alg. \ref{alg:CAKD}. In detail, the cross-architecture teacher-student framework is first constructed. The PCA projector and the GL projector are then embedded in the student network to map the student features into the teacher attention space and feature space. Subsequently, a cross-view robust training scheme is adopted to train the framework. The framework main body (\ie, $\mathbf{\Theta}^\mathrm{S}$, $\mathrm{Proj_1}(\cdot)$ and $\mathrm{Proj_2}(\cdot)$) and the multi-view adversarial discriminator $D(\cdot)$ are alternatively updated. After convergence, the modules $\mathrm{Proj_1}(\cdot)$, $\mathrm{Proj_2}(\cdot)$ and $D(\cdot)$ are removed and only the compact student network $\mathbf{\Theta}^\mathrm{S}$ is remained to carry out the inference phase.

\begin{algorithm}[!tb]
	\caption{The procedure of cross-architecture knowledge distillation.}
	\label{alg:CAKD}
	\LinesNumbered
	\KwIn{Database $\mathcal{D}^\mathrm{train}=\{\mathbf{x}^\mathrm{train},\mathbf{y}^\mathrm{train}\}$, $\mathbf{\Theta}^\mathrm{S}$, $\mathbf{\Theta}^\mathrm{T}$, $D(\cdot)$, $\mathrm{Proj_1}(\cdot)$, $\mathrm{Proj_2}(\cdot)$.}
	\BlankLine
	$e=0$; \\	
	Initialize $\mathbf{\Theta}^\mathrm{S}$, $\mathrm{Proj_1}(\cdot)$, $\mathrm{Proj_2}(\cdot)$ and $D(\cdot)$; \\
	\Repeat{done}{
			Compute the transformed features $\mathbf{h'}_{\mathrm{S}}$ and $\{Q_{\mathrm{S}}, K_{\mathrm{S}}, V_{\mathrm{S}}\}$ through $\mathrm{Proj_1}(\cdot)$ and $\mathrm{Proj_2}(\cdot)$, using Equation. (\ref{eqn:PCA}) and Equation (\ref{eqn:GL}); \\
			Update $\mathbf{\Theta}^\mathrm{S}$, $\mathrm{Proj_1}(\cdot)$ and $\mathrm{Proj_2}(\cdot)$ using Equation. (\ref{eqn:G_loss}); \\
			\uIf{$e \% 5 =0$}{
				Update $D(\cdot)$ using Equation. (\ref{eqn:D_loss}); }
			{end }  \\
			$ e = e + 1 $\;
		} 		
		Remove $\mathrm{Proj_1}(\cdot)$, $\mathrm{Proj_2}(\cdot)$ and $D(\cdot)$, and predict the label through $\mathbf{\Theta}^\mathrm{S}$ in inference phase;  \\
		return $\mathbf{\Theta}^\mathrm{S}$.
	\end{algorithm}

\section{Experiments}
\subsection{Settings}
\noindent\textbf{Databases and Networks.} We evaluate the proposed method on two databases: CIFAR~\cite{krizhevsky2009learning} and ImageNet~\cite{ILSVRC15}. The data are augmented using the same strategies as in the PyTorch official examples~\cite{paszke2017automatic}. 
For networks, we use the popular CNNs as the student network, including ResNets~\cite{he2016deep}, MobileNet v2~\cite{sandler2018mobilenetv2}, Xception~\cite{chollet2017xception} and EfficientNet~\cite{tan2019efficientnet}. 
The typical Transformers are applied as the teacher network, such as ViT~\cite{dosovitskiy2020image}, and Swin Transformer~\cite{liu2021swin}.
	
\noindent\textbf{Implementation Details.} 
We train all the networks from scratch. For CIFAR datasets, the total number of epochs is 200 with a standard batch size of 64. The learning rate is initialized as 0.1 and multiplied by 0.1 at epoch 100 and epoch 150. For ImageNet, the total number of epochs is 120 with a 256 batch size. The learning rate is initialized as 0.1 and multiplied by 0.1 at epoch 30, epoch 60 and epoch 90, respectively. A standard stochastic gradient descent (SGD) optimizer with $10^{-4}$ weight decay and 0.9 momentum is adopted. All the experiments are conducted on a platform with 8 Nvidia Tesla GPU cards and 96-core Intel(R) Xeon(R) Platinum 8163 CPU. In addition, every single setting is repeated 5 times with different random seeds on Pytorch.

\subsection{Performance Comparison}
We compare the performance of our method with 14 state-of-the-art knowledge distillation methods, including Logits~\cite{hinton2015distilling}, FitNet~\cite{romero2014fitnets}, AT~\cite{zagoruyko2016paying}, IRG~\cite{liu2019knowledge}, RKD~\cite{park2019relational}, CRD~\cite{tian2019contrastive}, OFD~\cite{heo2019comprehensive}, ReviewKD~\cite{chen2021distilling}, LONDON~\cite{shang2021lipschitz}, AFD~\cite{wang2019pay}, AB~\cite{heo2019knowledge}, FT~\cite{kim2018paraphrasing}, DeiT~\cite{touvron2021training} and MINILM~\cite{wang2020minilm}. Among them, Logits, FitNet, AT, IRG, RKD, CRD, OFD, ReviewKD and LONDON are CNN-based KD methods, and DeiT and MINILM are transformer-based KD methods. 
There exist few related works for the Transformer-CNN framework. Consequently, several CNN-based methods including logits, RKD and IRG are adopted for cross-architecture scenarios, since these methods do not rely on the CNN architectures.  
Besides, for a fair comparison, we select CNNs and Transformers with similar FLoating-point OPerations (FLOPs) or similar accuracy as the teacher network or the student network. 
	
\noindent\textbf{Evaluation on CIFAR.} 
Table \ref{tab:comparison_cifar} presents the KD results on CIFAR100. As shown in this table, three KD modes of the teacher-student frameworks, including CNN-CNN, Transformer-CNN and Transformer-Transformer, are evaluated.
It can be seen that the proposed method has the best performance among all the methods, including CNN-based KD methods and transformer-based methods. 
For the most commonly used CNN-CNN mode, 
the proposed cross-architecture KD method shows superior performance. It is because the CNN student learns complementary global information from the Transformer teacher. The performance gap is even larger (usually more than 1\%) when the Transformer teacher and the CNN teachers have similar FLOPs. Because under similar computation cost, Transformer teacher usually has higher accuracy than CNN teacher. 
For the Transformer-CNN mode, a higher performance gain (an average gain of 2.7\%) is obtained compared with the CNN-CNN methods. This indicates that existing KD methods do not take full advantage of the Transformer teacher, though they can be adopted to the cross-architecture scenario. 
In Transformer-Transformer mode, the proposed method results mostly surpass the Transformer-based KD results. Although the Xceptionx2 model is slightly inferior to the ViT-B/16 model, the performance gain of Xceptionx2 is higher than that of ViT-B/16. This indicates that cross-architecture KD can obtain higher promotion than the conventional homologous-architecture KD. 
Besides, in our cross-architecture framework, it is easier to adopt and accelerate the CNN student into practical application.

\noindent\textbf{Evaluation on ImageNet.}  
Experiments are conducted on ImageNet to further verify the generalization and effectiveness of the proposed method. As shown in Table \ref{tab:comparison_imagenet}, our method exhibits the best performance on ImageNet. Similar to the settings of CIFAR, two homologous-architecture modes including CNN-CNN and Transformer-Transformer and one cross-architecture mode, \ie, Transformer-CNN, are compared. Different from homologous-architecture methods, the proposed cross-architecture framework encourages the student to learn both local spatial features (with the original CNN model) and complementary global features (from the transformer teacher model). Consequently, the CNN student obtains higher performance. 
Especially, from Table \ref{tab:comparison_imagenet}, some CNNs (\eg, ResNet50x2-80.72\%) guided by Transformer even surpasses the Transformer with similar model computation (\eg, ViT-B/32-78.29\%), by more than 1.03\% accuracy. With hardware-friendly attributes, these improved CNNs are more potential for edge device applications.

\begin{table}[!htbp]
	\small
	\centering
	\setlength\tabcolsep{0.5pt}
	\setlength{\extrarowheight}{1.5pt}
	\caption{Performance comparison on CIFAR100. Note that ``x2'' denotes the channel number of this network is twice of the original ResNet's. And ``x3'' has the analogous meaning.}
	\vspace{-1em}
	\resizebox{1.01 \textwidth}{!}{\begin{tabular}{|c|c|c|c|c||c|c|c|c|} 
			\hline
			Mode
			& Teacher  &  Student                 &  Methods &  Test accuracy 
			& Teacher  &  Student                 &  Methods &  Test accuracy   \\ 
			\hline
			\multirow{11}*{{CNN$\rightarrow$CNN}}
			&\multirow{10}*{{\tabincell{c}{ResNet152x2 \\ \scriptsize(\textit{212.0 GFLOPs})}}}
			& \multirow{10}*{{\tabincell{c}{ResNet50 \\ \scriptsize(\textit{4.1 GFLOPs})}}}                              & Baseline\_T          & 91.03\%   
			&\multirow{10}*{{\tabincell{c}{ResNet101x3 \\ \scriptsize(\textit{205.0 GFLOPs})}}}
			& \multirow{10}*{{\tabincell{c}{ResNet50x2 \\ \scriptsize(\textit{15.9 GFLOPs})}}}                              & Baseline\_T           & 90.98\%              \\ 
			&&& Baseline\_S           & 85.02\%   
			&&& Baseline\_S           & 88.21\%       \\	
			\cdashline{4-5}   \cdashline{8-9}
			&&& Logits  & {86.53\%} 
			&&& Logits  & {89.07\%}\\ 
			&&& FitNet  & {85.37\%}
			&&& FitNet  & {88.51\%} \\ 
			&&& AT  & {86.41\%}
			&&& AT  & {89.18\%} \\ 
			&&& RKD  & {86.22\%}
			&&& RKD  & {89.39\%} \\ 
			&&& IRG  & {86.87\%}
			&&& IRG  & {89.89\%} \\ 
			&&& OFD  & {86.79\%}
			&&& OFD  & {89.62\%} \\ 
			&&& CRD  & {86.91\%}
			&&& CRD  & {89.94\%} \\ 
			&&& ReviewKD  & {87.03\%}
			&&& ReviewKD  & {90.04\%} \\ 
			&&& LONDON  & {87.16\%}
			&&& LONDON  & {89.98\%} \\ 
			\cdashline{2-9} 
			& ViT-B/16
			& ResNet50    & \textbf{Ours}      & 87.39\% 
			& ViT-B/16 
			& ResNet50x2    & \textbf{Ours}      &  90.33\%               \\ 
			& ViT-L/16
			& ResNet50    & \textbf{Ours}      & \textbf{88.09\%} 
			& ViT-L/16 
			& ResNet50x2    & \textbf{Ours}      & \textbf{90.97\%}                \\
			\hline \hline
			\multirow{18}*{{\tabincell{c}{Transformer \\ $\rightarrow$CNN}}}
			&\multirow{6}*{{\tabincell{c}{ViT-B/16 \\ \scriptsize(\textit{55.4 GFLOPs})}}}
			& \multirow{6}*{{\tabincell{c}{ResNet50 \\ \scriptsize(\textit{4.1 GFLOPs})}}}                              & Baseline\_T          & 90.92\%   
			&\multirow{6}*{{\tabincell{c}{ViT-L/16 \\ \scriptsize(\textit{190.7 GFLOPs})}}}
			& \multirow{6}*{{\tabincell{c}{ResNet50 \\ \scriptsize(\textit{4.1 GFLOPs})}}}                              & Baseline\_T           & 92.46\%              \\ 
			&&& Baseline\_S           & 85.02\%   
			&&& Baseline\_S           & 85.02\%       \\	
			\cdashline{4-5}   \cdashline{8-9}
			&&& Logits  & {86.42\%} 
			&&& Logits  & {86.69\%}\\ 
			&&& RKD  & {86.13\%}
			&&& RKD  & {86.73\%} \\ 
			&&& IRG  & {86.59\%}
			&&& IRG  & {86.91\%} \\ 
			\cdashline{4-5}   \cdashline{8-9}
			&&& \textbf{Ours}      & \textbf{87.39\%}
			&&& \textbf{Ours}      &  \textbf{88.09\%}               \\ 
			\cline{2-9}
			&\multirow{6}*{{\tabincell{c}{ViT-B/16 \\ \scriptsize(\textit{55.4 GFLOPs})}}}
			& \multirow{6}*{{\tabincell{c}{ResNet50x2 \\ \scriptsize(\textit{15.9 GFLOPs})}}}                              & Baseline\_T          & 90.92\%   
			&\multirow{6}*{{\tabincell{c}{ViT-L/16 \\ \scriptsize(\textit{190.7 GFLOPs})}}}
			& \multirow{6}*{{\tabincell{c}{ResNet50x2 \\ \scriptsize(\textit{15.9 GFLOPs})}}}                              & Baseline\_T           & 92.46\%              \\ 
			&&& Baseline\_S           & 88.21\%   
			&&& Baseline\_S           & 88.21\%       \\	
			\cdashline{4-5}   \cdashline{8-9}
			&&& Logits  & {88.86\%} 
			&&& Logits  & {89.28\%}\\ 
			&&& RKD  & {89.11\%}
			&&& RKD  & {89.51\%} \\ 
			&&& IRG  & {89.38\%}
			&&& IRG  & {89.68\%} \\ 
			\cdashline{4-5}   \cdashline{8-9}
			&&& \textbf{Ours}      & \textbf{90.33\%} 
			&&& \textbf{Ours}      &  \textbf{90.97\%}               \\ 
			\cline{2-9}
			&\multirow{6}*{{\tabincell{c}{Swin-L \\ \scriptsize(\textit{103.9 GFLOPs})}}}
			& \multirow{6}*{{\tabincell{c}{ResNet50 \\ \scriptsize(\textit{4.1 GFLOPs})}}}                              & Baseline\_T          & 93.78\%   
			&\multirow{6}*{{\tabincell{c}{Swin-L \\ \scriptsize(\textit{103.9 GFLOPs})}}}
			& \multirow{6}*{{\tabincell{c}{ResNet50x2 \\ \scriptsize(\textit{15.9 GFLOPs})}}}                              & Baseline\_T           & 93.78\%              \\ 
			&&& Baseline\_S           & 85.02\%   
			&&& Baseline\_S           & 88.21\%       \\	
			\cdashline{4-5}   \cdashline{8-9}
			&&& Logits  & {86.78\%} 
			&&& Logits  & {88.93\%}\\ 
			&&& RKD  & {86.91\%}
			&&& RKD  & {90.02\%} \\ 
			&&& IRG  & {87.06\%}
			&&& IRG  & {89.97\%} \\ 
			\cdashline{4-5}   \cdashline{8-9}
			&&& \textbf{Ours}      & \textbf{88.46\%} 
			&&& \textbf{Ours}      &  \textbf{91.21\%}               \\ 
			\hline \hline
			\multirow{15}*{{\tabincell{c}{Transformer \\ $\rightarrow$ \\ Transformer}}}
			&\multirow{6}*{{\tabincell{c}{ViT-L/16 \\ \scriptsize(\textit{190.7 GFLOPs})}}}
			& \multirow{6}*{{\tabincell{c}{ViT-B/16 \\ \scriptsize(\textit{55.4 GFLOPs})}}}                              & Baseline\_T          & 92.46\%   
			&\multirow{6}*{{\tabincell{c}{Swin-L \\ \scriptsize(\textit{103.9 GFLOPs})}}}
			& \multirow{6}*{{\tabincell{c}{ViT-B/16 \\ \scriptsize(\textit{55.4 GFLOPs})}}}                              & Baseline\_T           & 93.78\%              \\ 
			&&& Baseline\_S           & 90.92\%   
			&&& Baseline\_S           & 90.92\%       \\	
			\cdashline{4-5}   \cdashline{8-9}
			&&& Logits  & {91.45\%} 
			&&& Logits  & {91.74\%}\\ 
			&&& IRG  & {91.59\%}
			&&& IRG  & {91.88\%} \\ 
			&&& DeiT  & {91.57\%}
			&&& DeiT  & {91.91\%} \\ 
			&&& MINILM  & {91.44\%}
			&&& MINILM  & {91.75\%} \\ 
			\cdashline{2-9}  
			& \multirow{2}*{ViT-L/16}
			& \multirow{2}*{{\tabincell{c}{Xceptionx2 \\ \scriptsize(\textit{57.3G / 90.27\%})}}}    & \multirow{2}*{\textbf{Ours}}      & \multirow{2}*{91.15\%} 
			& \multirow{2}*{Swin-L} 
			& \multirow{2}*{{\tabincell{c}{Xceptionx2 \\ \scriptsize(\textit{57.3G / 90.27\%})}}}    & \multirow{2}*{\textbf{Ours}}      &  \multirow{2}*{91.36\%}               \\
			&&&&&&&&\\
			\cdashline{2-9}
			& ViT-L/16
			& ResNet101x3    & \textbf{Ours}      & \textbf{91.84\%} 
			& Swin-L 
			& ResNet101x3    & \textbf{Ours}      & \textbf{92.07\%}                \\
			\cline{2-9}
			&\multirow{6}*{{\tabincell{c}{ViT-L/16 \\ \scriptsize(\textit{190.7 GFLOPs})}}}
			& \multirow{6}*{{\tabincell{c}{ViT-B/32 \\ \scriptsize(\textit{13.8 GFLOPs})}}}                              & Baseline\_T          & 92.46\%   
			&\multirow{6}*{{\tabincell{c}{Swin-L \\ \scriptsize(\textit{103.9 GFLOPs})}}}
			& \multirow{6}*{{\tabincell{c}{ViT-B/32 \\ \scriptsize(\textit{13.8 GFLOPs})}}}                              & Baseline\_T           & 93.78\%              \\ 
			&&& Baseline\_S           & 89.46\%   
			&&& Baseline\_S           & 89.46\%       \\	
			\cdashline{4-5}   \cdashline{8-9}
			&&& Logits  & {90.22\%} 
			&&& Logits  & {90.59\%}\\ 
			&&& IRG  & {90.39\%}
			&&& IRG  & {90.95\%} \\ 
			&&& DeiT  & {90.40\%}
			&&& DeiT  & {90.99\%} \\ 
			&&& MINILM  & {90.26\%}
			&&& MINILM  & {90.62\%} \\ 
			\cdashline{2-9}  
			& \multirow{2}*{{\tabincell{c}{ViT-L/16 \\ \scriptsize(\textit{190.7 GFLOPs})}}}
			& \multirow{2}*{{\tabincell{c}{ResNet152 \\ \scriptsize(\textit{11.0 G / 89.57\%})}}}    
			& \multirow{2}*{\textbf{Ours}}      & \multirow{2}*{\textbf{90.66\%}} 
			& \multirow{2}*{{\tabincell{c}{Swin-L \\ \scriptsize(\textit{103.9 GFLOPs})}}} 
			& \multirow{2}*{{\tabincell{c}{ResNet152 \\ \scriptsize(\textit{11.0 G / 89.57\%})}}}    
			& \multirow{2}*{\textbf{Ours}}      &  \multirow{2}*{\textbf{91.20\%}}               \\
			&&&&&&&&\\
			\hline
	\end{tabular}  }
	\label{tab:comparison_cifar}
	\begin{tablenotes}
		\scriptsize
		\item[1] * Baseline\_T: Baseline model of the teacher network.
		\item[2] * Baseline\_S: Baseline model of the student network.
	\end{tablenotes}
\end{table}

\begin{table}[!tb]
	\small
	\centering
	\setlength\tabcolsep{0.5pt}
	\setlength{\extrarowheight}{2pt}
	\caption{Performance comparison on ImageNet.}
	\vspace{-1em}
	\resizebox{1.01 \textwidth}{!}{\begin{tabular}{|c|c|c|c|c||c|c|c|c|} 
			\hline
			\multirow{2}*{Mode}
			& \multirow{2}*{Teacher}  &  \multirow{2}*{Student}                 &  \multirow{2}*{Methods} &  \multirow{2}*{{\tabincell{c}{Test accuracy \\ Top1 / Top5}}} 
			& \multirow{2}*{Teacher}  &  \multirow{2}*{Student}                 &  \multirow{2}*{Methods} &  \multirow{2}*{{\tabincell{c}{Test accuracy \\ Top1 / Top5}}}   \\ 
			&&&&&&&&\\
			\hline
			\multirow{11}*{{CNN$\rightarrow$CNN}}
			&\multirow{10}*{{\tabincell{c}{ResNet152x2 \\ \scriptsize(\textit{212.0 GFLOPs})}}}
			& \multirow{10}*{{\tabincell{c}{ResNet50x2 \\ \scriptsize(\textit{15.9 GFLOPs})}}}                              & Baseline\_T          & 81.95 / 96.02   
			&\multirow{10}*{{\tabincell{c}{ResNet101x3 \\ \scriptsize(\textit{205.0 GFLOPs})}}}
			& \multirow{10}*{{\tabincell{c}{ResNet50x2 \\ \scriptsize(\textit{15.9 GFLOPs})}}}                              & Baseline\_T           & 82.03 / 96.06              \\ 
			&&& Baseline\_S           & 78.16 / 93.91   
			&&& Baseline\_S           & 78.16 / 93.91       \\	
			\cdashline{4-5}   \cdashline{8-9}
			&&& Logits  & {79.06 / 94.67} 
			&&& Logits  & {79.19 / 94.71}\\  
			&&& AT  & {79.01 / 94.66}
			&&& AT  & {78.92 / 94.63} \\ 
			&&& FT  & {79.12 / 94.69}
			&&& FT  & {79.11 / 94.69} \\
			&&& AB  & {78.93 / 94.62}
			&&& AB  & {79.01 / 94.65} \\ 
			&&& OFD  & {79.63 / 94.81}
			&&& OFD  & {79.55 / 94.79} \\ 
			&&& AFD  & {79.38 / 94.76}
			&&& AFD  & {79.45 / 94.78} \\ 
			&&& IRG  & {79.85 / 94.87}
			&&& IRG  & {79.75 / 94.84} \\ 
			&&& ReviewKD  & {80.12 / 94.99}
			&&& ReviewKD  & {80.08 / 94.97} \\ 
			&&& LONDON  & {80.09 / 94.97}
			&&& LONDON  & {80.15 / 95.01} \\ 
			\cdashline{2-9} 
			& ViT-B/16
			& ResNet50x2    & \textbf{Ours}      & 80.74 / 95.38 
			& ViT-B/16 
			& ResNet50x2    & \textbf{Ours}      &  80.72 / 95.38               \\ 
			& ViT-L/16
			& ResNet50x2    & \textbf{Ours}      & \textbf{80.92 / 95.43} 
			& ViT-L/16 
			& ResNet50x2    & \textbf{Ours}      & \textbf{81.01 / 95.46}                \\
			\hline \hline
			\multirow{18}*{{\tabincell{c}{Transformer \\ $\rightarrow$CNN}}}
			&\multirow{6}*{{\tabincell{c}{ViT-B/16 \\ \scriptsize(\textit{55.4 GFLOPs})}}}
			& \multirow{6}*{{\tabincell{c}{ResNet50 \\ \scriptsize(\textit{4.1 GFLOPs})}}}                              & Baseline\_T          & 82.17 / 96.11   
			&\multirow{6}*{{\tabincell{c}{ViT-L/16 \\ \scriptsize(\textit{190.7 GFLOPs})}}}
			& \multirow{6}*{{\tabincell{c}{ResNet50 \\ \scriptsize(\textit{4.1 GFLOPs})}}}                              & Baseline\_T           & 84.20 / 96.93              \\ 
			&&& Baseline\_S           & 76.28 / 93.03   
			&&& Baseline\_S           & 76.28 / 93.03       \\	
			\cdashline{4-5}   \cdashline{8-9}
			&&& Logits  & {77.02 / 93.40} 
			&&& Logits  & {77.45 / 93.57}\\ 
			&&& RKD  & {77.27 / 93.50}
			&&& RKD  & {77.82 / 93.75} \\ 
			&&& IRG  & {77.39 / 93.55}
			&&& IRG  & {77.75 / 93.71} \\ 
			\cdashline{4-5}   \cdashline{8-9}
			&&& \textbf{Ours}      & \textbf{78.34 / 94.06}
			&&& \textbf{Ours}      &  \textbf{78.85 / 94.31}               \\ 
			\cline{2-9}
			&\multirow{6}*{{\tabincell{c}{ViT-B/16 \\ \scriptsize(\textit{55.4 GFLOPs})}}}
			& \multirow{6}*{{\tabincell{c}{ResNet50x2 \\ \scriptsize(\textit{15.9 GFLOPs})}}}                              & Baseline\_T          & 82.17 / 96.11   
			&\multirow{6}*{{\tabincell{c}{ViT-L/16 \\ \scriptsize(\textit{190.7 GFLOPs})}}}
			& \multirow{6}*{{\tabincell{c}{ResNet50x2 \\ \scriptsize(\textit{15.9 GFLOPs})}}}                              & Baseline\_T           & 84.20 / 96.93              \\ 
			&&& Baseline\_S           & 78.16 / 93.91   
			&&& Baseline\_S           & 78.16 / 93.91       \\	
			\cdashline{4-5}   \cdashline{8-9}
			&&& Logits  & {79.02 / 94.62} 
			&&& Logits  & {79.31 / 94.72}\\ 
			&&& RKD  & {79.68 / 94.82}
			&&& RKD  & {79.78 / 94.85} \\ 
			&&& IRG  & {79.60 / 94.79}
			&&& IRG  & {79.83 / 94.88} \\ 
			\cdashline{4-5}   \cdashline{8-9}
			&&& \textbf{Ours}      & \textbf{80.72 / 95.38} 
			&&& \textbf{Ours}      &  \textbf{81.01 / 95.46}               \\
			\cline{2-9}
			&\multirow{6}*{{\tabincell{c}{Swin-L \\ \scriptsize(\textit{103.9 GFLOPs})}}}
			& \multirow{6}*{{\tabincell{c}{ResNet50 \\ \scriptsize(\textit{4.1 GFLOPs})}}}                              & Baseline\_T          & 87.32 / 98.21   
			&\multirow{6}*{{\tabincell{c}{Swin-L \\ \scriptsize(\textit{103.9 GFLOPs})}}}
			& \multirow{6}*{{\tabincell{c}{ResNet50x2 \\ \scriptsize(\textit{15.9 GFLOPs})}}}                              & Baseline\_T           & 87.32 / 98.21              \\ 
			&&& Baseline\_S           & 76.28 / 93.03   
			&&& Baseline\_S           & 78.16 / 93.91       \\	
			\cdashline{4-5}   \cdashline{8-9}
			&&& Logits  & {77.60 / 93.64} 
			&&& Logits  & {79.68 / 94.83}\\ 
			&&& RKD  & {77.85 / 93.76}
			&&& RKD  & {79.92 / 94.92} \\ 
			&&& IRG  & {77.89 / 93.79}
			&&& IRG  & {80.10 / 94.99} \\ 
			\cdashline{4-5}   \cdashline{8-9}
			&&& \textbf{Ours}      & \textbf{78.96 / 94.42} 
			&&& \textbf{Ours}      &  \textbf{81.39 / 95.64}               \\ 
			\hline \hline
			\multirow{15}*{{\tabincell{c}{Transformer \\ $\rightarrow$ \\ Transformer}}}
			&\multirow{6}*{{\tabincell{c}{ViT-L/16 \\ \scriptsize(\textit{190.7 GFLOPs})}}}
			& \multirow{6}*{{\tabincell{c}{ViT-B/16 \\ \scriptsize(\textit{55.4 GFLOPs})}}}                              & Baseline\_T          & 84.20 / 96.93   
			&\multirow{6}*{{\tabincell{c}{Swin-L \\ \scriptsize(\textit{103.9 GFLOPs})}}}
			& \multirow{6}*{{\tabincell{c}{ViT-B/16 \\ \scriptsize(\textit{55.4 GFLOPs})}}}                              & Baseline\_T           & 87.32 / 98.21              \\ 
			&&& Baseline\_S           & 82.17 / 96.11   
			&&& Baseline\_S           & 82.17 / 96.11       \\	
			\cdashline{4-5}   \cdashline{8-9}
			&&& Logits  & {83.18 / 96.55} 
			&&& Logits  & {83.49 / 96.65}\\ 
			&&& IRG  & {83.27 / 96.59}
			&&& IRG  & {83.60 / 96.69} \\ 
			&&& DeiT  & {83.38 / 96.63}
			&&& DeiT  & {83.71 / 96.72} \\ 
			&&& MINILM  & {83.17 / 96.55}
			&&& MINILM  & {83.55 / 96.65} \\ 
			\cdashline{2-9}  
			& \multirow{2}*{ViT-L/16}
			& \multirow{2}*{{\tabincell{c}{Xceptionx2 \\ \scriptsize(\textit{80.37\% / 95.24\%})}}}    & \multirow{2}*{\textbf{Ours}}      & \multirow{2}*{82.56 / 96.34} 
			& \multirow{2}*{Swin-L} 
			& \multirow{2}*{{\tabincell{c}{Xceptionx2 \\ \scriptsize(\textit{80.37\% / 95.24\%})}}}    & \multirow{2}*{\textbf{Ours}}      & \multirow{2}*{82.98 / 96.45}                \\
			&&&&&&&&\\
			\cdashline{2-9} 
			& ViT-L/16
			& ResNet152x2    & \textbf{Ours}      & \textbf{83.62 / 96.74} 
			& Swin-L 
			& ResNet101x3    & \textbf{Ours}      &  \textbf{84.37 / 96.97}               \\
			\cline{2-9}
			&\multirow{6}*{{\tabincell{c}{ViT-L/16 \\ \scriptsize(\textit{190.7 GFLOPs})}}}
			& \multirow{6}*{{\tabincell{c}{ViT-B/32 \\ \scriptsize(\textit{13.8 GFLOPs})}}}                              & Baseline\_T          & 84.20 / 96.93   
			&\multirow{6}*{{\tabincell{c}{Swin-L \\ \scriptsize(\textit{103.9 GFLOPs})}}}
			& \multirow{6}*{{\tabincell{c}{ViT-B/32 \\ \scriptsize(\textit{13.8 GFLOPs})}}}                              & Baseline\_T           & 87.32 / 98.21              \\ 
			&&& Baseline\_S           & 78.29 / 94.08   
			&&& Baseline\_S           & 78.29 / 94.08       \\	
			\cdashline{4-5}   \cdashline{8-9}
			&&& Logits  & {79.40 / 94.76} 
			&&& Logits  & {79.30 / 94.73}\\ 
			&&& IRG  & {79.20 / 94.64}
			&&& IRG  & {79.10 / 94.60} \\ 
			&&& DeiT  & {79.37 / 94.75}
			&&& DeiT  & {79.27 / 94.71} \\ 
			&&& MINILM  & {79.29 / 94.70}
			&&& MINILM  & {79.19 / 94.67} \\ 
			\cdashline{2-9}  
			& \multirow{2}*{ViT-L/16}
			& \multirow{2}*{{\tabincell{c}{ResNet152 \\ \scriptsize(\textit{78.31\% / 94.05\%})}}}    & \multirow{2}*{\textbf{Ours}}      & \multirow{2}*{\textbf{80.47 / 95.29}} 
			& \multirow{2}*{Swin-L} 
			& \multirow{2}*{{\tabincell{c}{ResNet152 \\ \scriptsize(\textit{78.31\% / 94.05\%})}}}    & \multirow{2}*{\textbf{Ours}}      &  \multirow{2}*{\textbf{81.09 / 95.52}}               \\
			&&&&&&&&\\
			\hline
	\end{tabular}  }
	\label{tab:comparison_imagenet}
\end{table}

\subsection{Ablation Study} 
\textbf{(1) Different teacher-student pairs.} 
In order to verify the generalization of the proposed method, we evaluate it with different cross-architecture teacher-student pairs in Table \ref{tab:diff_pairs}. It can be observed that our cross-architecture method obtains significant performance promotion across different teacher-student pairs, compared with the baseline. In addition, the accuracies of the student continue increasing as the teacher’s performance becomes better. At this end, Transformer can be an excellent teacher since it usually obtains better performance with similar FLOPs compared with a CNN network. Using Transformer to guide the learning of a CNN student can be a potential direction.

\begin{table}[!tb]
	\small
	\centering
	\setlength{\extrarowheight}{0.5pt}
	\caption{Performance results of different teacher-student pairs on ImageNet. Note that the brackets behind the networks report the FLOPs of the networks.}
	\vspace{-1em}
	\resizebox{.96\textwidth}{!}{
		\begin{tabular}{|c|c|c|c|c|c||c|c|} 
			\hline
			\multirow{2}*{Teacher}   &  \multirow{2}*{Student}
			&\multicolumn{2}{c|}{Teacher accuracy}
			&\multicolumn{2}{c||}{Student accuracy}
			& \multicolumn{2}{c|}{Ours accuracy}       \\ 
			\cline{3-8} 
			&&Top1  &Top5 &Top1 &Top5 &Top1 &Top5 \\
			\hline	
			ViT-B/16 \scriptsize(\textit{55.4G})& \multirow{5}*{{\tabincell{c}{ResNet50 \\ \scriptsize(\textit{4.1 GFLOPs})}}} & 82.17\% & 96.11\% & 76.28\% & 93.03\% & 78.34\% & 94.06\% \\
			ViT-L/16 \scriptsize(\textit{190.7G}) &  & 84.20\% & 96.93\% & 76.28\% & 93.03\% & 78.85\% & 94.31\% \\
			DeiT-B \scriptsize(\textit{55.4G}) &  & 83.12\% & 96.52\% & 76.28\% & 93.03\% & 78.53\% & 94.13\% \\
			Swin-B \scriptsize(\textit{15.4G}) &  & 86.38\% & 98.01\% & 76.28\% & 93.03\% & 78.87\% & 94.29\% \\
			Swin-L \scriptsize(\textit{103.9G}) &  & 87.32\% & 98.21\% & 76.28\% & 93.03\% & 78.96\% & 94.42\% \\
			\cdashline{1-8}
			ViT-B/16 & \multirow{5}*{{\tabincell{c}{ResNet18 \\ \scriptsize(\textit{1.9 GFLOPs})}}} & 82.17\% & 96.11\% & 69.76\% & 89.08\% & 71.73\% & 90.41\% \\
			ViT-L/16 &  & 84.20\% & 96.93\% & 69.76\% & 89.08\% & 72.02\% & 90.52\% \\
			DeiT-B &  & 83.12\% & 96.52\% & 69.76\% & 89.08\% & 71.85\% & 90.45\% \\
			Swin-B &  & 86.38\% & 98.01\% & 69.76\% & 89.08\% & 72.01\% & 90.52\% \\
			Swin-L &  & 87.32\% & 98.21\% & 69.76\% & 89.08\% & 72.09\% & 90.57\% \\
			\cdashline{1-8}
			ViT-B/16 & \multirow{5}*{{\tabincell{c}{MobileNetV2 \\ \scriptsize(\textit{0.3 GFLOPs})}}} & 82.17\% & 96.11\% & 71.88\% & 90.29\% & 73.34\% & 91.01\% \\
			ViT-L/16 &  & 84.20\% & 96.93\% & 71.88\% & 90.29\% & 73.52\% & 91.18\% \\
			DeiT-B &  & 83.12\% & 96.52\% & 71.88\% & 90.29\% & 73.40\% & 91.06\% \\
			Swin-B &  & 86.38\% & 98.01\% & 71.88\% & 90.29\% & 73.56\% & 91.21\% \\
			Swin-L &  & 87.32\% & 98.21\% & 71.88\% & 90.29\% & 73.66\% & 91.25\% \\
			\cdashline{1-8}
			ViT-B/16 & \multirow{5}*{{\tabincell{c}{EfficientNetB0 \\ \scriptsize(\textit{1.6 GFLOPs})}}} & 82.17\% & 96.11\% & 77.69\% & 93.53\% & 79.23\% & 94.50\% \\
			ViT-L/16 &  & 84.20\% & 96.93\% & 77.69\% & 93.53\% & 79.34\% & 94.54\% \\
			DeiT-B &  & 83.12\% & 96.52\% & 77.69\% & 93.53\% & 79.30\% & 94.52\% \\
			Swin-B &  & 86.38\% & 98.01\% & 77.69\% & 93.53\% & 79.38\% & 94.55\% \\
			Swin-L &  & 87.32\% & 98.21\% & 77.69\% & 93.53\% & 79.52\% & 94.60\% \\
			\hline	
	\end{tabular}  }
	\vspace{-2em}
	\label{tab:diff_pairs}
\end{table}

\begin{figure}[!htbp]
	\begin{center}
		\vspace{-.2em}
		\includegraphics[width=.93\linewidth]{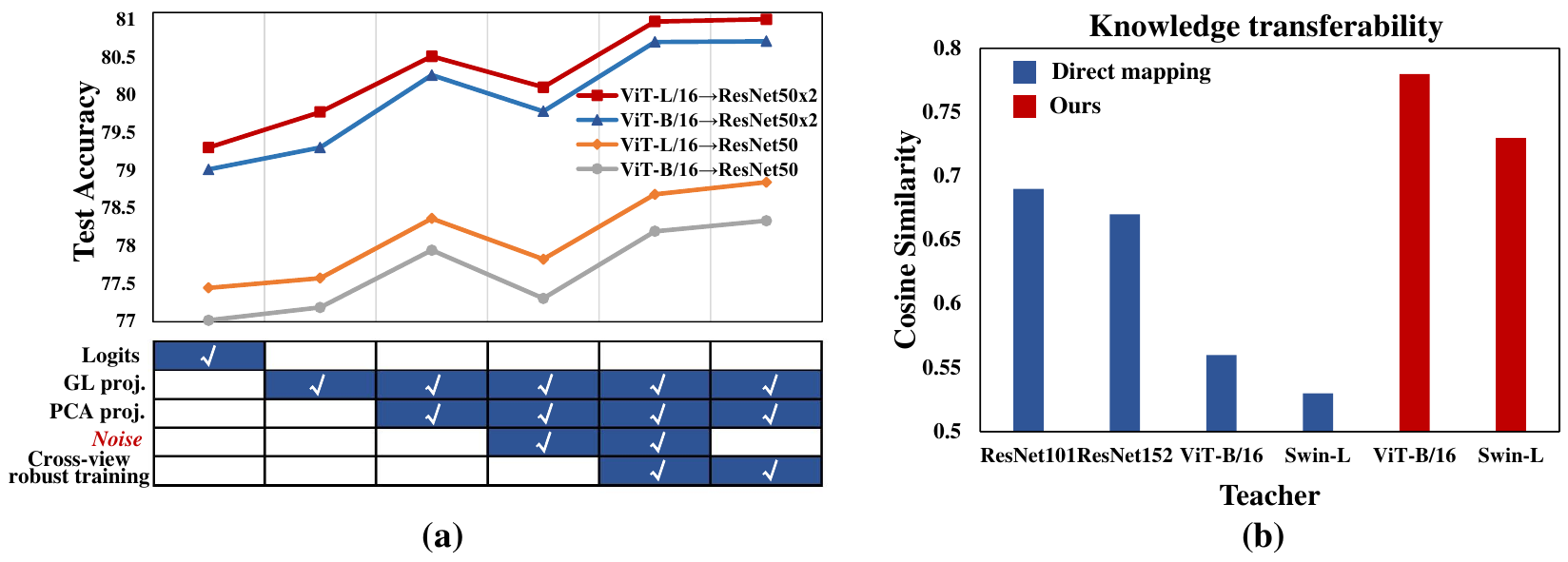}
	\end{center}
	\vspace{-2.2em}
	\caption{(a) Performance of each component in the proposed method. (b) The cosine similarities between the features from different models. The student network is ResNet50. Among these blue bars, the features are mapped into the same dimension with the teacher features by a linear projector. All the results are obtained on ImageNet.}
	\vspace{-2.0em}
	\label{fig:ablation}
\end{figure}

\textbf{(2) Effectiveness of the proposed projector.} We analyze the effectiveness of the proposed PCA projector and GL projector. Experimental results on ImageNet in Figure \ref{fig:ablation}-(a) show great performance gain when the two projectors are involved during the KD procedure. It indicates that PCA and GL projectors significantly improve the quality of the CNN feature, though they are removed during the inference phase. We further evaluate the transferability after adding these two projectors in Figure \ref{fig:ablation}-(b). The cosine similarity is increased by a large margin and is even higher than that of the homologous-architecture. Therefore, it is possible to increase the knowledge transferability between Transformer and CNN by carefully designed KD methods.

\textbf{(3) Effectiveness of the cross-view robust training.} 
As reported in Figure \ref{fig:ablation}-(a), for regular evaluation without noise, student networks obtain 0.2\%-0.4\% top-1 accuracy gain on ImageNet with the cross-view robust training scheme. To further verify its effectiveness, we also report the results for noisy evaluation, where the validation dataset is augmented differently from the training augmentation. Under this protocol, the top-1 accuracy gain after adding the cross-view robust training scheme is enlarged to more than 1.0\%. It demonstrates that the proposed robust training scheme enhances the noise robustness of the student network.

\textbf{(4) Applications on other tasks.} 
The proposed cross-architecture KD method also performs well on other tasks. As shown in Tab. \ref{tab:other_tasks}, our method is evaluated on three visual tasks including object detection~\cite{ren2015faster}, instance segmentation~\cite{he2017mask} and face anti-spoofing~\cite{zhang2020celeba}. 

For detection and segmentation, 
we follow the recent protocol of the COCO database~\cite{lin2014microsoft} and report 
average precision (AP). Note that AP in segmentation is computed using mask intersection over union (IoU). 
The proposed method shows superiority compared with the conventional KD method in Tab. \ref{tab:other_tasks}. For the conventional KD method Logits, the performance of the cross-architecture mode is even worse than the performance of the homologous-architecture mode. This further manifests that our method effectively solves the mismatching problem of cross-architecture KD. 
In addition, for face anti-spoofing, which is a binary classification task, we adopt ResNet18, Inception-v3 and ResNext26 as the backbones of the student. Equal Error Rate (EER) is reported as the evaluation metric. And the experiments are conducted on CelebA-Spoof~\cite{zhang2020celeba}, which is one of the largest datasets for face anti-Spoofing. It is worth mentioning that there exist few useful information of class correlation on the binary classification task. Hence, conventional KD method Logits has marginal enhancement on the student. In contrast, the proposed method also obtains a satisfactory performance from Tab. \ref{tab:other_tasks}.
It is interesting to notice that, though the proposed method is designed for the classification task, it has good generalization when it is directly applied to other tasks such as detection and segmentation.

\begin{table}[!tb]
	\small
	\centering
	\setlength{\extrarowheight}{0.5pt}
	\caption{Evaluation on other visual tasks, including object detection, instance segmentation and face anti-spoofing.}
	\vspace{-1em}
	\resizebox{.76\textwidth}{!}{
		\begin{tabular}{|c|c|c|c|c|c|c|} 
			\hline
			{Task (Dataset)}   &  {Teacher backbone}
			&{Student backbone}
			& {Method}                 & {AP} 
			&{$\Delta$AP}    \\ 
			\hline	
			\multirow{12}*{\tabincell{c}{\textbf{Object} \\ \textbf{Detection} \\ (COCO)}} 
			& 	{$--$} 
			&\multirow{4}*{{\normalsize ResNet50}}
			& Baseline          & 34.5            & 0                \\ 
			&ResNet152x2&& Logits          & 35.0            & 0.5             \\
			&ViT-L/16&& Logits          & 34.9            & 0.4             \\ 
			&ViT-L/16&& \textbf{Ours}          & \textbf{35.5}            & \textbf{1.0}            \\ \cdashline{2-6}
			& {$--$}
			&\multirow{4}*{{\normalsize ResNet101}}
			& Baseline          & 37.1            & 0                \\ 
			&ResNet152x2&& Logits          & 37.7            & 0.6             \\
			&ViT-L/16&& Logits          & 37.4            & 0.3             \\ 
			&ViT-L/16&& \textbf{Ours}          & \textbf{38.1}            & \textbf{1.0}      \\     \cdashline{2-6}
			& {$--$}
			& \multirow{4}*{{\normalsize ResNeXt101}}
			& Baseline          & 39.2            & 0                \\ 
			&ResNet152x2&& Logits          & 39.8           & 0.6             \\
			&ViT-L/16&& Logits          & 39.6            & 0.4             \\ 
			&ViT-L/16&& \textbf{Ours}          & \textbf{40.3}            & \textbf{1.1}      \\ 
			\hline \hline
			{Task (Dataset)}   &  {Teacher backbone}
			&{Student backbone}
			& {Method}                 & {AP} 
			&{$\Delta$AP}    \\ 
			\hline	
			\multirow{12}*{\tabincell{c}{\textbf{Instance} \\ \textbf{Segmentation} \\ (COCO)}} 
			& 	$--$ 
			&\multirow{4}*{{\normalsize ResNet50}}
			& Baseline          & 32.6            & 0                \\ 
			&ResNet152x2&& Logits          & 33.3           & 0.7             \\
			&ViT-L/16&& Logits          & 33.1            & 0.5             \\ 
			&ViT-L/16&& \textbf{Ours}          & \textbf{33.6}            & \textbf{1.0}      \\  \cdashline{2-6}
			& $--$
			&\multirow{4}*{{\normalsize ResNet101}}
			& Baseline          & 33.9            & 0                \\ 
			&ResNet152x2&& Logits          & 34.5           & 0.6             \\
			&ViT-L/16&& Logits          & 34.2            & 0.3             \\ 
			&ViT-L/16&& \textbf{Ours}          & \textbf{34.8}            & \textbf{0.9}      \\  \cdashline{2-6}
			& $--$
			& \multirow{4}*{{\normalsize ResNeXt101}}
			& Baseline          & 35.1            & 0                \\ 
			&ResNet152x2&& Logits          & 35.5           & 0.4             \\
			&ViT-L/16&& Logits          & 35.3            & 0.2             \\ 
			&ViT-L/16&& \textbf{Ours}          & \textbf{35.9}            & \textbf{0.8}      \\ 
			\hline \hline
			{Task (Dataset)}   &  {Teacher backbone}
			&{Student backbone}
			& {Method}                 & {EER} 
			&{$-\Delta$EER}    \\ 
			\hline	
			\multirow{12}*{\tabincell{c}{\textbf{Face} \\ \textbf{Anti-Spoofing} \\ (CelebA-Spoof)}} 
			& 	$--$
			&\multirow{4}*{{\normalsize ResNet18}}
			& Baseline          & 1.6            & 0                \\ 
			&ResNet152x2&& Logits          & 1.6           & 0             \\
			&ViT-L/16&& Logits          & 1.6            & 0             \\ 
			&ViT-L/16&& \textbf{Ours}          & \textbf{1.3}            & \textbf{0.3}      \\  \cdashline{2-6}
			& $--$
			&\multirow{3}*{{\normalsize Inception-v3}}
			& Baseline          & 1.4            & 0                \\ 
			&ResNet152x2&& Logits          & 1.3           & 0.1             \\
			&ViT-L/16&& Logits          & 1.4            & 0             \\ 
			&ViT-L/16&& \textbf{Ours}          & \textbf{1.1}            & \textbf{0.3}      \\  \cdashline{2-6}
			& $--$
			&\multirow{3}*{{\normalsize ResNeXt26}}
			& Baseline          & 1.3            & 0                \\ 
			&ResNet152x2&& Logits          & 1.3           & 0             \\
			&ViT-L/16&& Logits          & 1.3            & 0             \\ 
			&ViT-L/16&& \textbf{Ours}          & \textbf{0.9}            & \textbf{0.4}      \\ 
			\hline
	\end{tabular}  }
	\vspace{-2em}
	\label{tab:other_tasks}
\end{table}

\section{Conclusions}
In this paper, 
a novel cross-architecture knowledge distillation method is proposed. In particular, 
two projectors including a partially cross attention (PCA) projector and a group-wise Linear (GL) projector are presented 
The two projectors promote the knowledge transferability from teacher to student. In order to further improve the robustness and stability of the framework, a multi-view robust training scheme is proposed. 
Extensive experimental results show that our method outperforms 14 state-of-the-arts on both large-scale datasets and small-scale datasets.

\subsubsection{Acknowledgements} This work was supported by the National Key Research and Development Program of China (Grant No. 2020AAA0106800), the National Natural Science Foundation of China (No. 62192785, Grant No.61902401, No. 61972071, No. U1936204, No. 62122086, No. 62036011, No. 62192782 and No. 61721004), the Beijing Natural Science Foundation No. M22005, the CAS Key Research Program of Frontier Sciences (Grant No. QYZDJ-SSW-JSC040). The work of Bing Li was also supported by the Youth Innovation Promotion Association, CAS.

%
%
%
\bibliographystyle{splncs}
\bibliography{egbib}

\begin{thebibliography}{10}

\bibitem{cheng2017survey}
Cheng, Y., Wang, D., Zhou, P., Zhang, T.:
\newblock A survey of model compression and acceleration for deep neural
  networks.
\newblock arXiv preprint arXiv:1710.09282 (2017)

\bibitem{tan2018survey}
Tan, C., Sun, F., Kong, T., Zhang, W., Yang, C., Liu, C.:
\newblock A survey on deep transfer learning.
\newblock In: International conference on artificial neural networks, Springer
  (2018)  270--279

\bibitem{dosovitskiy2020image}
Dosovitskiy, A., Beyer, L., Kolesnikov, A., Weissenborn, D., Zhai, X.,
  Unterthiner, T., Dehghani, M., Minderer, M., Heigold, G., Gelly, S.,  et~al.:
\newblock An image is worth 16x16 words: Transformers for image recognition at
  scale.
\newblock arXiv preprint arXiv:2010.11929 (2020)

\bibitem{carion2020end}
Carion, N., Massa, F., Synnaeve, G., Usunier, N., Kirillov, A., Zagoruyko, S.:
\newblock End-to-end object detection with transformers.
\newblock In: European conference on computer vision, Springer (2020)  213--229

\bibitem{cuda_2007}
Nvidia:
\newblock Cuda.
\newblock In: https://developer.nvidia.com/cuda-zone, Nvidia (2007)

\bibitem{tensorrt_2022}
Nvidia:
\newblock Tensorrt.
\newblock In: https://developer.nvidia.com/tensorrt, Nvidia (2022)

\bibitem{ncnn_2017}
Tencent:
\newblock Ncnn.
\newblock In: https://github.com/Tencent/ncnn, Tencent (2017)

\bibitem{ILSVRC15}
Russakovsky, O., Deng, J., Su, H., Krause, J., Satheesh, S., Ma, S., Huang, Z.,
  Karpathy, A., Khosla, A., Bernstein, M., Berg, A.C., Fei-Fei, L.:
\newblock {ImageNet Large Scale Visual Recognition Challenge}.
\newblock International Journal of Computer Vision \textbf{115} (2015)
  211--252

\bibitem{krizhevsky2009learning}
Krizhevsky, A., Hinton, G.:
\newblock Learning multiple layers of features from tiny images.
\newblock Technical report, Citeseer (2009)

\bibitem{hinton2015distilling}
Hinton, G., Vinyals, O., Dean, J.:
\newblock Distilling the knowledge in a neural network.
\newblock arXiv preprint arXiv:1503.02531 (2015)

\bibitem{ba2013deep}
Ba, L.J., Caruana, R.:
\newblock Do deep nets really need to be deep?
\newblock arXiv preprint arXiv:1312.6184 (2013)

\bibitem{zagoruyko2016paying}
Zagoruyko, S., Komodakis, N.:
\newblock Paying more attention to attention: Improving the performance of
  convolutional neural networks via attention transfer.
\newblock arXiv preprint arXiv:1612.03928 (2016)

\bibitem{romero2014fitnets}
Romero, A., Ballas, N., Kahou, S.E., Chassang, A., Gatta, C., Bengio, Y.:
\newblock Fitnets: Hints for thin deep nets.
\newblock arXiv preprint arXiv:1412.6550 (2014)

\bibitem{heo2019comprehensive}
Heo, B., Kim, J., Yun, S., Park, H., Kwak, N., Choi, J.Y.:
\newblock A comprehensive overhaul of feature distillation.
\newblock In: Proceedings of the IEEE/CVF International Conference on Computer
  Vision. (2019)  1921--1930

\bibitem{huang2017like}
Huang, Z., Wang, N.:
\newblock Like what you like: Knowledge distill via neuron selectivity
  transfer.
\newblock arXiv preprint arXiv:1707.01219 (2017)

\bibitem{yim2017gift}
Yim, J., Joo, D., Bae, J., Kim, J.:
\newblock A gift from knowledge distillation: Fast optimization, network
  minimization and transfer learning.
\newblock In: Proceedings of the IEEE Conference on Computer Vision and Pattern
  Recognition. (2017)  4133--4141

\bibitem{liu2019knowledge}
Liu, Y., Cao, J., Li, B., Yuan, C., Hu, W., Li, Y., Duan, Y.:
\newblock Knowledge distillation via instance relationship graph.
\newblock In: Proceedings of the IEEE/CVF Conference on Computer Vision and
  Pattern Recognition. (2019)  7096--7104

\bibitem{song2022spot}
Song, J., Chen, Y., Ye, J., Song, M.:
\newblock Spot-adaptive knowledge distillation.
\newblock IEEE Transactions on Image Processing \textbf{31} (2022)  3359--3370

\bibitem{song2021tree}
Song, J., Zhang, H., Wang, X., Xue, M., Chen, Y., Sun, L., Tao, D., Song, M.:
\newblock Tree-like decision distillation.
\newblock In: Proceedings of the IEEE/CVF Conference on Computer Vision and
  Pattern Recognition. (2021)  13488--13497

\bibitem{touvron2021training}
Touvron, H., Cord, M., Douze, M., Massa, F., Sablayrolles, A., J{\'e}gou, H.:
\newblock Training data-efficient image transformers \& distillation through
  attention.
\newblock In: International Conference on Machine Learning, PMLR (2021)
  10347--10357

\bibitem{wang2020minilm}
Wang, W., Wei, F., Dong, L., Bao, H., Yang, N., Zhou, M.:
\newblock Minilm: Deep self-attention distillation for task-agnostic
  compression of pre-trained transformers.
\newblock arXiv preprint arXiv:2002.10957 (2020)

\bibitem{aguilar2020knowledge}
Aguilar, G., Ling, Y., Zhang, Y., Yao, B., Fan, X., Guo, C.:
\newblock Knowledge distillation from internal representations.
\newblock In: Proceedings of the AAAI Conference on Artificial Intelligence.
  Volume~34. (2020)  7350--7357

\bibitem{paszke2017automatic}
Paszke, A., Gross, S., Chintala, S., Chanan, G., Yang, E., DeVito, Z., Lin, Z.,
  Desmaison, A., Antiga, L., Lerer, A.:
\newblock Automatic differentiation in pytorch.
\newblock In: Advances in Neural Information Processing Systems Workshop.
  (2017)

\bibitem{he2016deep}
He, K., Zhang, X., Ren, S., Sun, J.:
\newblock Deep residual learning for image recognition.
\newblock In: Proceedings of the IEEE Conference on Computer Vision and Pattern
  Recognition. (2016)  770--778

\bibitem{sandler2018mobilenetv2}
Sandler, M., Howard, A., Zhu, M., Zhmoginov, A., Chen, L.C.:
\newblock Mobilenetv2: Inverted residuals and linear bottlenecks.
\newblock In: Proceedings of the IEEE Conference on Computer Vision and Pattern
  Recognition. (2018)  4510--4520

\bibitem{chollet2017xception}
Chollet, F.:
\newblock Xception: Deep learning with depthwise separable convolutions.
\newblock In: Proceedings of the IEEE conference on computer vision and pattern
  recognition. (2017)  1251--1258

\bibitem{tan2019efficientnet}
Tan, M., Le, Q.:
\newblock Efficientnet: Rethinking model scaling for convolutional neural
  networks.
\newblock In: International conference on machine learning, PMLR (2019)
  6105--6114

\bibitem{liu2021swin}
Liu, Z., Lin, Y., Cao, Y., Hu, H., Wei, Y., Zhang, Z., Lin, S., Guo, B.:
\newblock Swin transformer: Hierarchical vision transformer using shifted
  windows.
\newblock In: Proceedings of the IEEE/CVF International Conference on Computer
  Vision. (2021)  10012--10022

\bibitem{park2019relational}
Park, W., Kim, D., Lu, Y., Cho, M.:
\newblock Relational knowledge distillation.
\newblock In: Proceedings of the IEEE/CVF Conference on Computer Vision and
  Pattern Recognition. (2019)  3967--3976

\bibitem{tian2019contrastive}
Tian, Y., Krishnan, D., Isola, P.:
\newblock Contrastive representation distillation.
\newblock arXiv preprint arXiv:1910.10699 (2019)

\bibitem{chen2021distilling}
Chen, P., Liu, S., Zhao, H., Jia, J.:
\newblock Distilling knowledge via knowledge review.
\newblock In: Proceedings of the IEEE/CVF Conference on Computer Vision and
  Pattern Recognition. (2021)  5008--5017

\bibitem{shang2021lipschitz}
Shang, Y., Duan, B., Zong, Z., Nie, L., Yan, Y.:
\newblock Lipschitz continuity guided knowledge distillation.
\newblock In: Proceedings of the IEEE/CVF International Conference on Computer
  Vision. (2021)  10675--10684

\bibitem{wang2019pay}
Wang, K., Gao, X., Zhao, Y., Li, X., Dou, D., Xu, C.Z.:
\newblock Pay attention to features, transfer learn faster cnns.
\newblock In: International conference on learning representations. (2019)

\bibitem{heo2019knowledge}
Heo, B., Lee, M., Yun, S., Choi, J.Y.:
\newblock Knowledge transfer via distillation of activation boundaries formed
  by hidden neurons.
\newblock In: Proceedings of the AAAI Conference on Artificial Intelligence.
  Volume~33. (2019)  3779--3787

\bibitem{kim2018paraphrasing}
Kim, J., Park, S., Kwak, N.:
\newblock Paraphrasing complex network: Network compression via factor
  transfer.
\newblock Advances in neural information processing systems \textbf{31} (2018)

\bibitem{ren2015faster}
Ren, S., He, K., Girshick, R., Sun, J.:
\newblock Faster r-cnn: Towards real-time object detection with region proposal
  networks.
\newblock Advances in neural information processing systems \textbf{28} (2015)
  91--99

\bibitem{he2017mask}
He, K., Gkioxari, G., Doll{\'a}r, P., Girshick, R.:
\newblock Mask r-cnn.
\newblock In: Proceedings of the IEEE international conference on computer
  vision. (2017)  2961--2969

\bibitem{zhang2020celeba}
Zhang, Y., Yin, Z., Li, Y., Yin, G., Yan, J., Shao, J., Liu, Z.:
\newblock Celeba-spoof: Large-scale face anti-spoofing dataset with rich
  annotations.
\newblock In: European Conference on Computer Vision, Springer (2020)  70--85

\bibitem{lin2014microsoft}
Lin, T.Y., Maire, M., Belongie, S., Hays, J., Perona, P., Ramanan, D.,
  Doll{\'a}r, P., Zitnick, C.L.:
\newblock Microsoft coco: Common objects in context.
\newblock In: European conference on computer vision, Springer (2014)  740--755

\end{thebibliography}
%
%
%
%
%

\end{document}